\documentclass[journal]{IEEEtran}
\ifCLASSINFOpdf
\else
\fi
\usepackage{mathtools}
\usepackage{algorithm}
\usepackage{algpseudocode}
\usepackage{graphicx}
\usepackage{booktabs,chemformula}
\DeclareGraphicsExtensions{.pdf,.png,.jpg}
\usepackage{hyperref}
\hypersetup{
    colorlinks=true,
    linkcolor=blue,
    filecolor=cyan,
    urlcolor=magenta,
}

\begin{document}
%
\title{Key Points Estimation and Point Instance Segmentation Approach for Lane Detection}
%
%
%

\author{Yeongmin Ko,~\IEEEmembership{Student Member,~IEEE,}
       Younkwan Lee,~\IEEEmembership{Student Member,~IEEE,}
       Shoaib Azam,~\IEEEmembership{Student Member,~IEEE},
       Farzeen Munir,~\IEEEmembership{Senior Member,~IEEE},
       Moongu Jeon*,~\IEEEmembership{Senior Member,~IEEE}, and
       Witold Pedrycz,~\IEEEmembership{Fellow,~IEEE}
\thanks{This work was partly supported by Institute of Information communications Technology Planning Evaluation (IITP) grant funded by the Korea Government (MSIT) (No. 2014-3-00077, Development of Global Multi-target Tracking and Event Prediction Techniques Based on Real-time Large-Scale Video Analysis), the National Research Foundation of Korea (NRF) grant funded by the Korea Government (MSIT) (No. 2019R1A2C2087489), and GIST Research Institute(GRI) grant funded by the GIST in 2019.}
\thanks{Y. Ko, Y. Lee, S. Azam, F. Munir and M. Jeon are with the School of Electrical Engineering and Computer Science, Gwangju Institute of Science and Technology (GIST),
Gwangju, 61005, South Korea (e-mail: \{koyeongmin, brightyoun, shoaibazam, farzeen.munir, mgjeon\}@gist.ac.kr).}
\thanks{W. Pedrycz is with the Department of Electrical and Computer Engineering, University of Alberta, Edmonton, AB T6R 2V4, Canada, with the Department of Electrical and Computer Engineering, Faculty of Engineering, King Abdulaziz University, Jeddah 21589, Saudi Arabia, and also with the Systems Research Institute, Polish Academy of Sciences, Warsaw 01-447, Poland (email: wpedrycz@ualberta.ca).}
\thanks{Moongu Jeon is the corresponding author (e-mail: mgjeon@gist.ac.kr)}
}

%
%

\markboth{}%
{Shell \MakeLowercase{\textit{et al.}}: Bare Demo of IEEEtran.cls for IEEE Journals}
%



\maketitle
\begin{abstract}
Perception techniques for autonomous driving should be adaptive to various environments. In the case of traffic line detection, an essential perception module, many condition should be considered, such as number of traffic lines and computing power of the target system. To address these problems, in this paper, we propose a traffic line detection method called Point Instance Network (PINet); the method is based on the key points estimation and instance segmentation approach. The PINet includes several stacked hourglass networks that are trained simultaneously. Therefore the size of the trained models can be chosen according to the computing power of the target environment. We cast a clustering problem of the predicted key points as an instance segmentation problem; the PINet can be trained regardless of the number of the traffic lines. The PINet achieves competitive accuracy and false positive on the TuSimple and Culane datasets, popular public datasets for lane detection. Our code is available at
\url{https://github.com/koyeongmin/PINet_new}
\end{abstract}


%
\IEEEpeerreviewmaketitle

\section{Introduction}
%
%
%
%
\IEEEPARstart{F}{ully} autonomous driving requires understanding the environment around vehicles. Various perception modules are fused for this understanding, and many pattern recognition and computer vision techniques are applied for these perception modules \cite{lee2019snider}, \cite{munir2019localization}. Lane detection, which can localize the drivable area on a road, is a major perception technique. There are many ways to recognize lanes, but most techniques utilize traffic line detection \cite{wang2004lane}, \cite{kim2008robust} or road region segmentation \cite{LI20161}, \cite{wang2019self}. In this paper, we focus on traffic line detection for recognizing lanes. Fig. 1 shows the purpose of our proposed method, which predicts exact key points of lanes from input RGB images and, using embedding features extracted by the proposed network, distinguishes key points into individual instances. In addition, the proposed network is trained end-to-end, and the network size can be modified according to the computing power of the target system without any change of the network architecture or additional training. 
%

\begin{figure}
    \centering
    \includegraphics[width=0.48\textwidth]{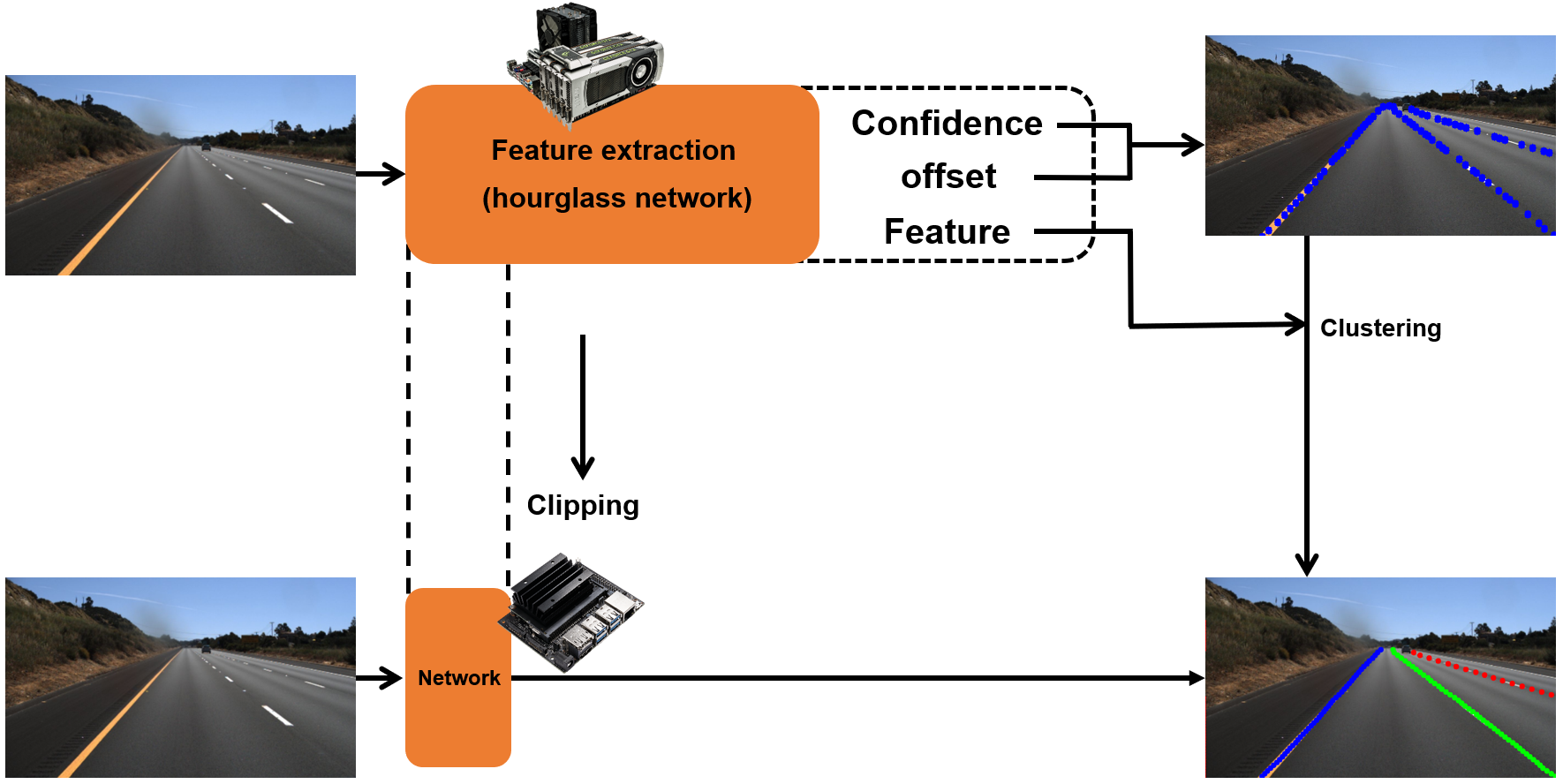}
    \caption{System overview. The proposed framework predicts key points on traffic lines and distinguishes individual instances regardless of the number of traffic lines. In addition, if user wants to run the trained model on a system with weak computing power, like an embedded board, the network can be clipped and transferred without additional training.}
\end{figure}

Most traditional methods of traffic line detection extract low-level traffic line features using various hand-craft features like color \cite{1}, \cite{2}, or edges \cite{wang1998lane}, \cite{lee2018robust}. These low-level features can be combined using a Hough transform \cite{duda1972use}, \cite{9146381} or Kalman filter \cite{borkar2009robust}; the combined features generate traffic line segment information. These methods are simple and can be adapted to various environments without significant modification. Still, the performance of these methods depends on condition of the testing environment such as lighting and occlusion.

\begin{figure*}
    \centering
    \includegraphics[width=4.8in,height=2.5in]{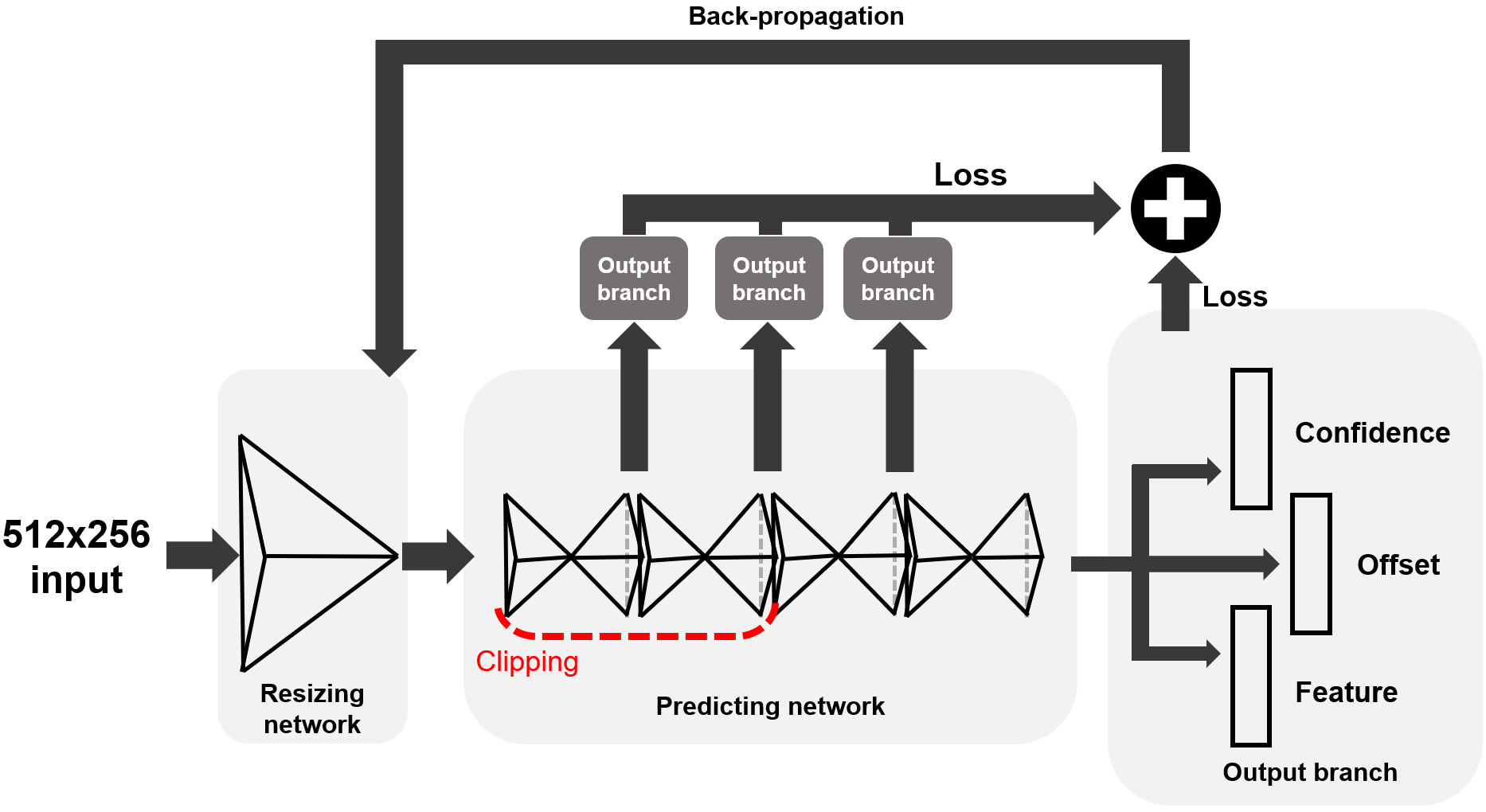}
    \caption{Proposed framework with three main parts. $512\times256$ size input data is compressed by the resizing network; the compressed input is fed to the predicting network, which includes four hourglass modules. Three output branches are applied at the ends of each hourglass block; they predict confidence, offset, and embedding feature. The loss function can be calculated from the outputs of each hourglass block. By clipping several hourglass modules, required computing resources can be adjusted.
}
\end{figure*}

Deep learning methods show outstanding performance for complex scenes. Among deep learning methods, Convolutional Neural Network (CNN) methods are primarily applied for feature extraction in computer vision \cite{krizhevsky2012imagenet}, \cite{he2017mask}. Semantic segmentation methods \cite{badrinarayanan2017segnet}, \cite{paszke2016enet}, \cite{ronneberger2015u}, the major research area in computer vision, are frequently applied to traffic line detection problems to make inferences about shapes and locations \cite{yang2019improved}, \cite{pan2018spatial}, \cite{van2019end}, \cite{zou2019robust}. Some methods use multi-class approaches to distinguish individual traffic line instances. Therefore, even though these methods can achieve outstanding performance, they can only be applied to scenes that consist of fixed numbers of traffic lines. As a solution to this problem, instance segmentation methods are applied to distinguish individual instance. These semantic segmentation based traffic line detection methods require some post-processing to estimate the exact location values of the predicted traffic lines. Avoiding this post-processing of the semantic segmentation approach, several other methods directly predict traffic line location \cite{chen2019pointlanenet}, \cite{8624563}.

The existing methods have certain limitations. The semantic segmentation methods require the labeling or pre-processing at the pixel level for training, which is cumbersome. These methods also predict many unnecessary points because semantic segmentation generates classified pixel images with sizes identical to the given input image, even though only a few points are required to recognize traffic lines. In addition, existing methods are not adaptive to various environments according to available computing power. To apply them to light systems like embedded boards, the entire architecture should be modified and trained again.

To overcome these limitations, our proposed method uses a deep learning model inspired by a stacked hourglass network to predict a few key points on traffic lines. The stacked hourglass network \cite{newell2016stacked} is usually applied in key points estimation fields such as pose estimation \cite{yang2017learning} and object detection \cite{9010985}, \cite{zhou2019bottom}. Using sequence of down-sampling and up-sampling, the stacked hourglass network can extract information about various scales. Because the stacked hourglass network includes several hourglass modules that are trained by the same loss function, we can simultaneously obtain various models that have different parameter sizes by clipping some bays from the whole structure. Using the simple method inspired by point cloud instance segmentation, each key point is distinguished into individual instance \cite{wang2018sgpn}.

Camera-based traffic line detection has been actively developed, and many state-of-the-art methods \cite{hou2019learning}, \cite{8624563} are almost completely effective for public data sets. However, some methods have higher rates of false positive. False negatives, traffic lines that the module fails to detect, do not suddenly change the control values, and correct control values can be predicted from other detected traffic lines or previous results. However, false positives can lead to severe risks; incorrect identification of traffic lines by the module can cause rapid changes of the control values.

In summary, Fig. 2 shows our proposed framework for traffic line detection. It has three output branches and predicts the exact location and instance features of points on traffic lines. More details are introduced in section III. These are the primary contributions of this study:

\begin{itemize}

\item Using the key points estimation approach, we propose a novel method for traffic line detection. It produces a more compact size prediction output than those of other semantic segmentation-based methods.

\item The framework consists of several hourglass modules, and so we can obtain various models that have different sizes by simple clipping because each hourglass module is trained simultaneously using the same loss function.

\item The proposed method can be applied to various scenes that include any orientation of traffic lines, such as vertical or horizontal traffic lines, and arbitrary numbers of traffic lines.

\item The proposed method has lower false positives and the noteworthy accuracy performance. It guarantees the stability of the autonomous driving car.

\end{itemize}

\section{Related Work}

\subsection{Traffic Line Detection}
Lane detection is an important research area in autonomous driving. Lane detection modules recognize drivable areas on roads from input data. Traffic line detection is considered a main method for lane detection. Traffic line detection usually localizes line markings that distinguish drivable areas on roads. Especially regarding RGB images as input data, various hand-crafted features have been proposed to detect traffic lines \cite{deusch2012random}, \cite{yim2003three}, \cite{borkar2011novel}, \cite{8454897}, \cite{6719504}. However, these methods show limitations in complex scenarios.

\begin{figure*}
    \centering
    \includegraphics[width=4.8in,height=2.5in]{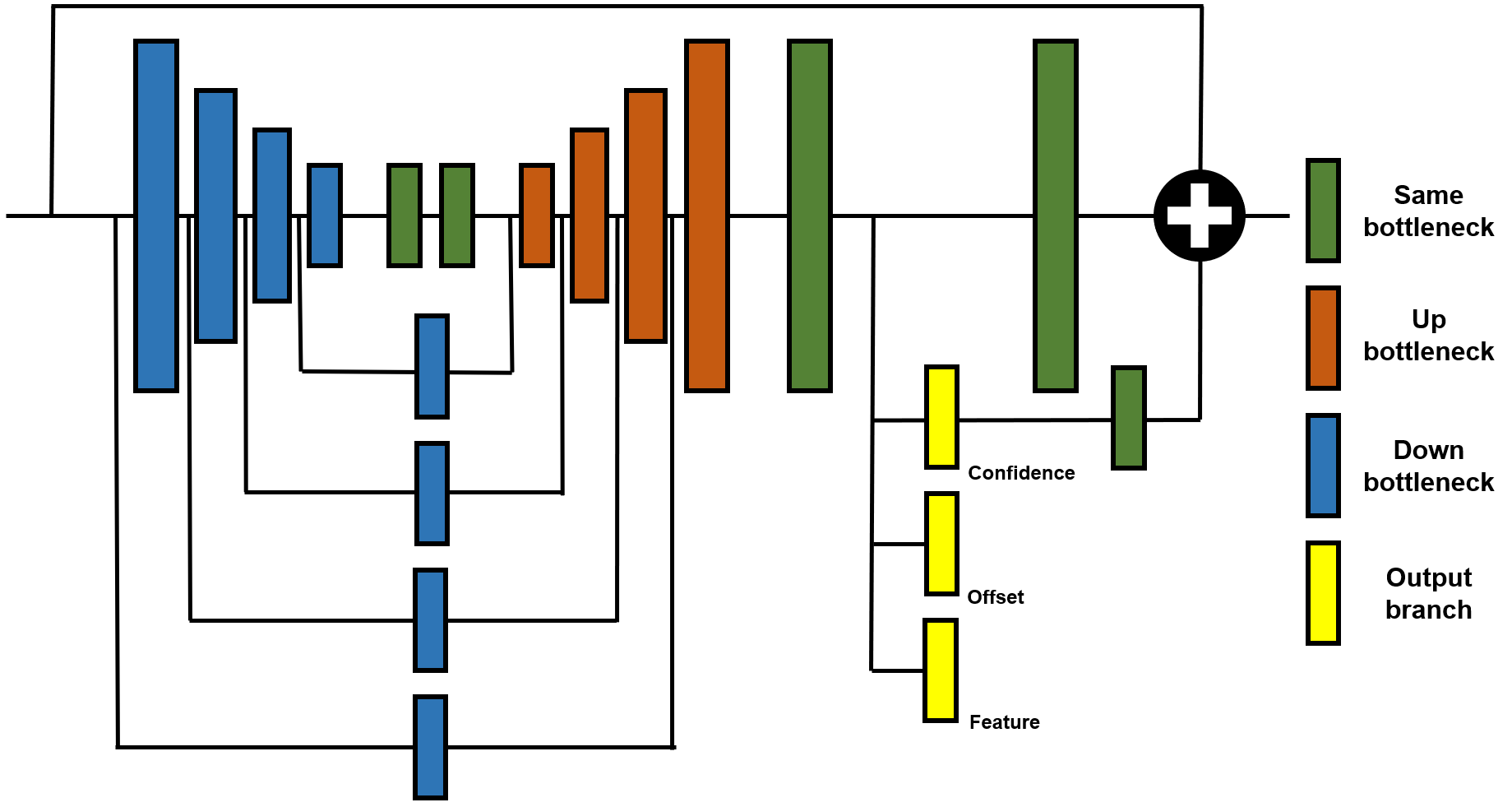}
    \caption{Details of hourglass block consisting three types of bottle-neck layers: same bottle-necks, down bottle-necks, and up bottle-necks. Output branches are applied at ends of hourglass layers; confidence output is forwarded to the next block.}
\end{figure*}

Recently, deep learning has become a dominant method in computer vision research. Semantic segmentation \cite{badrinarayanan2017segnet}, \cite{paszke2016enet}, \cite{ronneberger2015u} \cite{choi2020adfnet} is a major topic in perception research; it can classify pixels of the input image into individual class. Generative methods \cite{goodfellow2014generative}, \cite{zhu2017unpaired} can also perform a similar function. Therefore, semantic segmentation methods and generative methods are suitable for expressing complex shapes of lines. \cite{pan2018spatial}, \cite{hou2019learning}, \cite{lo2019multi}, and \cite{ghafoorian2018gan} show applications of semantic segmentation and the generative model for traffic line detection. Some methods use multi-class approaches to distinguish each instance; however, multi-class approaches can classify only fixed numbers of instances. Instance segmentation approaches are proposed as solutions to this limitation. Neven \textit{et al.} \cite{neven2018towards} attempted to solve this problem of multi-class approaches with instance segmentation. Their proposed LaneNet has a shared encoder and two decoders. One of these decoders performs binary lane segmentation; the other predicts embedding features for instance segmentation.

Although semantic segmentation methods can predict lines that have complex shapes, during training and testing they require pixel-level labeled data and post-processing to extract exact points on lines. Some direct methods \cite{chen2019pointlanenet}, \cite{8624563} directly generate exact points on lines. \cite{chen2019pointlanenet} predicts exact starting and terminal points,  and x-axis values of the fixed y-axis values for each traffic line. \cite{8624563} presents the Line Proposal Unit (LPU) inspired by the Region Proposal Network (RPN) of Faster R-CNN \cite{ren2015faster}. LPU predicts horizontal offsets for fixed y-axis values along certain pre-defined line proposals.

These approaches, the semantic segmentation method, the generative method, and the direct method, produce many unnecessary output values. In semantic segmentation and generative method, not all pixels are required to recognize traffic lines; an exact line can be predicted from a few key points. Direct methods also have certain unnecessary predictions like the length, starting points, and terminal points of the given target traffic lines that are unknown.

\subsection{Key Points Estimation}

Key points estimation techniques predict from input images certain important points called key points. Human pose estimation \cite{yang2017learning} is a major research topic in the key points estimation area. Stacked hourglass networks \cite{newell2016stacked} consists of several hourglass modules that are trained simultaneously. The hourglass module can transfer various scales' information to deeper layers, helping the whole network obtain both global and local features. Because of this property, an hourglass network is frequently utilized to detect centers or corners of objects in the object detection area. Not only network architecture or loss function but also refinement methods adapted to existing networks are developed for key point estimation. \cite{li2019rethinking} suggests a feature aggregation and coarse-to-fine supervision method that can be applied to other multi-stage methods. \cite{moon2019posefix} proposes the refinement network that improves the results of other existing models. In this paper, these refinement methods are not applied to indicate performance of our proposed framework; however, they can be applied to improve the performance.

\section{Method}

For lane detection, we train a neural network that consists of several hourglass modules. The network, which we will refer to as the Point Instance Network (PINet), generates points on lanes and distinguishes predicted points into individual instance. To achieve these tasks, our proposed neural network includes three output branches, a confidence branch, offset branch, and embedding branch. The confidence and offset branches predict exact points of traffic lines; loss functions inspired from YOLO \cite{redmon2016you} are applied. The embedding branch generates the embedding features of each predicted point; the embedding feature is fed to the clustering process to distinguish each instance. The loss function of the embedding branch is inspired by an instance segmentation method. The Similarity Group Proposal Network (SPGN) \cite{wang2018sgpn}, an instance segmentation frameworks for 3D point cloud, introduces a simple technique and a loss function for instance segmentation. Based on the contents proposed by SPGN, we design a loss function fitting to discriminate each instance of the predicted traffic lines. Section II-A introduces details of the main architecture; Section II-B consists of details about the loss function; and Section II-C shows the implementation in detail.

\subsection{Architecture}
Fig. 2 shows the proposed framework of the network. Input RGB image size is $512 \times 256$; it is fed to the resizing network. This image is compressed to a smaller size ($64 \times 32$) by the sequence of convolution layers in the resizing network; the output of the resizing network is fed to the predicting network. An arbitrary number of hourglass modules can be included in the predicting network; four hourglass modules are used in this study. All hourglass modules are trained simultaneously by the same loss function. After the training step, user can choose how many hourglass modules to use according to the computing power, without any additional training. The following sections provide details about each network.

\begin{figure}
    \centering
    \includegraphics[width=0.35\textwidth]{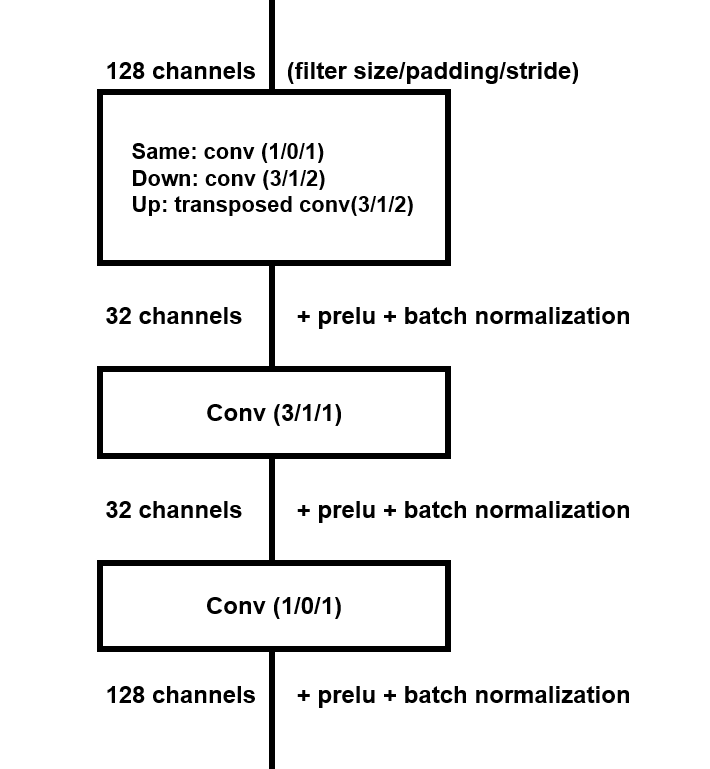}
    \caption{Details of bottle-neck. The three kinds of bottle-neck have different first layers according to their purposes.}
\end{figure}

\subsubsection{Resizing Network}
The resizing network reduces the input image's size to save memory and inference time. First, the input RGB image size is $512 \times 256$. This network consists of three convolution layers. All convolution layers are applied with filter size $3 \times 3$, stride 2, and padding size 1. Prelu \cite{he2015delving} and batch normalization \cite{batchnorm} are utilized after each convolution layer. Finally, this network generates resized output with $64 \times 32$ size. Table I shows details of the constituent layers. 

\begin{table}[ht]
    \caption{Details of resizing network}
    \begin{center}
        \begin{tabular}{|c|c|c|}
            \hline
            \textbf{Layer} & \textbf{Size/Stride} & \textbf{Output size}\\
            \hline
            \hline
            Input data &  & 3*512*256\\
            Conv+Prelu+bn & 3/2 & 32*256*128\\
            Conv+Prelu+bn & 3/2 & 64*128*64\\
            Conv+Prelu+bn & 3/2 & 128*64*32\\
            \hline
        \end{tabular}
    \end{center}
\end{table}

\subsubsection{Predicting Network}
The resizing network output is fed to the prediction part, which will be described in this section. This part predicts the exact points on the traffic lines and the embedding features for instance segmentation. This network consists of several hourglass modules, each including an encoder, decoder, and three output branches, as shown in Fig. 3. Some skip-connections transfer the information of the various scales to deeper layers. Each colored block in Fig. 3 is a bottle-neck module; these bottle-neck modules are described in Fig. 4. There are three kinds of bottle-neck: same, down, and up bottle-necks. The same bottle-neck generates output that has the same size as the input. The down bottle-neck is applied for down-sampling in the encoder; the first layer of the down bottle-neck is replaced by a convolution layer with filter size 3, stride 2, and padding 1. The transposed convolution layer with filter size 3, stride 2, and padding 1 is applied for the up bottle-neck in the up-sampling layers. Each output branch has three convolution layers, and generates a $64 \times 32$ grid. Confidence values about key point existence, offset, and embedding feature of each cell in the output grid are predicted by the output branches. Table II shows details of the predicting network. Because a deeper network has better performance \cite{newell2016stacked}, it can act as a teacher network. Therefore, using knowledge distillation techniques, we can expect better performance for clipped short networks. The channel of each output branch is different (confidence: 1, offset: 2, embedding: 4), and the corresponding loss function is applied according to the goal of each output branch.

\begin{table}[ht]
    \caption{Details of predicting network}
    \begin{center}
        \begin{tabular}{|c|c|c|c|}
            \hline
            & \textbf{Layer} & \textbf{Size/Stride} & \textbf{Output size}\\
            \hline
            \hline
            & Input data &  & 128*64*32\\
            \hline
            \hline
            Encoder & Bottle-neck(down) &  & 128*32*16\\
            & Bottle-neck(down) &  & 128*16*8\\
            & Bottle-neck(down) &  & 128*8*4\\
            & Bottle-neck(down) &  & 128*4*2\\
            & Bottle-neck &  & 128*4*2\\
            (Distillation layer)& Bottle-neck &  & 128*4*2\\
            & Bottle-neck &  & 128*4*2\\
            & Bottle-neck &  & 128*4*2\\
            \hline
            \hline
            Decoder & Bottle-neck(up) &  & 128*8*4\\
            & Bottle-neck(up) &  & 128*16*8\\
            & Bottle-neck(up) &  & 128*32*16\\
            & Bottle-neck(up) &  & 128*64*32\\
            \hline
            \hline
            Output branch & Conv+Prelu+bn & 3/1 & 64*64*32\\
            & Conv+Prelu+bn & 3/1 & 32*64*32\\
            & Conv & 1/1 & C*64*32\\
            \hline
        \end{tabular}
    \end{center}
\end{table}

\subsection{Loss Function}
For training, four loss functions are applied to each output branch of the hourglass networks. The following sections provide details of each loss function. As in Table II, the output branch generates a $64 \time 32$ grid, and each cell in the output grid consist of the predicted values of 7 channels, including the confidence value (1 channel), offset (2 channel) value, and embedding feature (4 channel). Confidence value determines whether or not key points of the traffic line exist; offset value localizes the exact position of the key points predicted by the confidence value, and the embedding feature is utilized to distinguish key points into individual instance. Therefore, three loss functions, except for the distillation loss function, are applied to each cell of the output grid. The distillation loss function to distillate the knowledge of the teacher network is adapted to the distillation layer of each encoder, as shown in Table II. Details of each predicted value and feature are included by the following sections.


\subsubsection{Confidence Loss}
The confidence output branch predicts the confidence value of each cell. If a key point is present in the cell, the confidence value is close to 1, if not, it is 0. The output of the confidence branch has 1 channel, and it is fed to the next hourglass module. The confidence loss consists of two parts, existence loss and non-existence loss. The existence loss is applied to cells that include key points; the non-existence loss is utilized to reduce the confidence value of each background cell. The non-existence loss is computed at cells that predict confidence values higher than 0.01. Because cells away from key points converge rapidly, this technique helps the training concentrate on cells closer to the key points. The following shows the loss function of the confidence branch:
\begin{equation}
    \begin{aligned}
        L_{exist} = \frac{1}{N_{e}} \sum_{c_c \in G_{e}}{(c_c^*-c_c)}^2 ,
    \end{aligned}
\end{equation}
\begin{equation}
    \begin{aligned}
        L_{non\_exist} = \, &\frac{1}{N_{n}} \sum_{\substack{c_c \in G_{n} \\ c_c > 0.01}}{(c_c^*-c_c)^2} 
        \\
        & + 0.00001 \cdot \sum_{c_c \in G_{n}}{c_c^2} ,
    \end{aligned}
\end{equation}
where $N_{e}$ denotes the number of cells that include key points, $N_{n}$ denotes the number of cells that do not include any key points, $G_e$ denotes a set of cells that consist of key points, $G_n$ denotes a set of cells that consist of points, $c_c$ denotes the predicted value of each cell in the confidence output branch, and $c_c^*$ denotes the ground-truth value. The ground truth value of the cell that has key point is 1; otherwise it is 0. At inference time, if the confidence value is bigger than the pre-defined threshold, we consider that a key point exists at the cell. The second term of $L_{non\_exist}$ is a regularization term.

\subsubsection{Offset Loss}
From the offset branch, PINet predicts the exact location of the key points for each output cell. The output of each cell has a value between 0 and 1; this value indicates the position related to the corresponding cell. In this paper, a cell is matched to 8 pixels of the input image. For example, if the predicted offset value is 0.5, the real position of the key point is 4 pixels away from the edge of the cell. The offset branch has two channels for predicting the x-axis and y-axis offsets. Equation 2 shows the loss function:

\begin{equation}
    \begin{aligned}
        L_{offset} = \, &\frac{1}{N_{e}} \sum_{c_x \in G_{e}}{(c_x^*-c_x)}^2
        \\
        &+ \frac{1}{N_{e}} \sum_{c_y \in G_{e}}(c_y^*-c_y)^2 .
    \end{aligned}
\end{equation}

Because the ground truth does not exist at cells that include no key points, these cells are ignored when the offset loss is calculated.

\subsubsection{Embedding Feature Loss}
The loss function of this branch is inspired by SGPN, a 3D points cloud instance segmentation method \cite{wang2018sgpn}. The branch is trained to make the embedding feature of each cell closer if the embedding features are the same in this instance. Equations 3 and 4 show the loss function of the feature branch:

\begin{equation}
    \begin{aligned}
        & L_{feature} = \frac{1}{N_{e}^2} \sum_{i}^{N_e} \sum_{j}^{N_e} l(i,j) ,
        \\
        l(i,j) = & 
        \begin{cases}
            ||F_i - F_j||_2 & \text{ if } I_{ij} = 1  \\ 
            max(0, K - ||F_i - F_j||_2) & \text{ if } I_{ij} = 0
        \end{cases} ,       
    \end{aligned}
\end{equation}

where $F_i$ denotes the predicted embedding feature of a cell i, $I_{ij}$ indicates whether cell i and cell j are same instance or not, and K is a constant such that $K>0$. If $I_{ij} = 1$, the cells are the same instance, and if $I_{ij} = 0$, these cells are different instances. When the network is trained, the loss function makes features closer when each cell belongs to the same instance; it distributes features when cells belong to different instances. We can distinguish key points into individual instance using the simple distance-based clustering technique. In this study, if embedding features of certain predicted key points are within a certain distance, we consider that they are the same instance. The feature size is set at 4 in this study, but this size is observed to have no major effect on the performance.

\subsubsection{Distillation Loss}
According to Newell \textit{et al.} \cite{newell2016stacked}, better performance is observed when more hourglass modules are stacked. Therefore, the deepest hourglass module can be a teacher network, and we expect that clipped short networks that are lighter than the teacher network will show better performance if a knowledge distillation method is applied. Zagoruyko \& Komodakis \cite{zagoruyko2016paying} proposed a simple knowledge distillation method that can be applied to the CNN model. This method allows a student network to imitate a teacher network; Hou \textit{et al.} \cite{hou2019learning} show that the method can improve the performance of the whole framework. Equation 5. shows the loss function for distillation:

\begin{equation}
    \begin{aligned}
        L_{distillation} & = \sum_{m}^{M}{D(F(A_M) - F(A_m))} ,
        \\
        F(A_M) & = S(G(A_m)), \;\; S : spatial \; softmax ,
        \\
        G(A_m) & = \sum_{i=1}^{C}{\mid A_{mi} \mid^2}, \;\; G : R^{C \times H \times W} \rightarrow R^{ H \times W} ,
    \end{aligned}
\end{equation}

where $D$ denotes the sum of square, $A_m$ denotes the distillation layer output at the m-th hourglass modules, as shown in Table II, $M$ denotes the number of hourglass modules, $A_mi$ denotes the i-th channel of $A_m$, and all operators like sum, power, and absolute value ($\mid \cdot \mid$) are elementwise.

The total loss $L_{total}$ is equal to the weighted sum of the above four loss terms, and the whole network is trained using an end-to-end procedure with the following total loss:
\begin{equation}
    \begin{aligned}
        L_{total}= & \; \gamma_e L_{exist} + \gamma_n L_{non-exist} + \gamma_o L_{offset}
        \\
        & + \gamma_f L_{feature} + \gamma_d L_{distillation} .
    \end{aligned}
\end{equation}
In the training step, we set $\gamma_o$ to 0.2, $\gamma_f$ to 0.5, and $\gamma_d$ to 0.1. $\gamma_e$ and $\gamma_n$ are described at Section IV. The proposed loss function is adapted to the output branch of each hourglass module; this helps the whole network to be trained stably.

\begin{figure}
    \centering
    \includegraphics[width=0.35\textwidth]{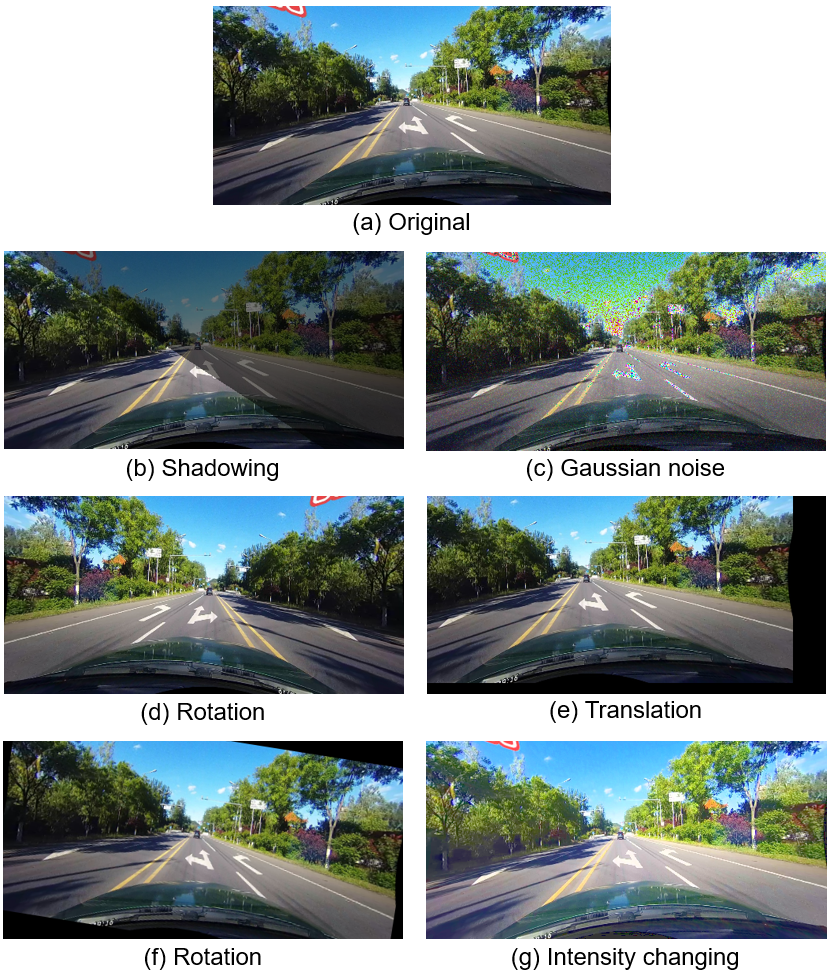}
    \caption{Data augmentation methods. (a) is the original image, and (b), (c), (d), (e), (f), and (g) show examples of the applied data augmentation methods.}
\end{figure}

\begin{table*}[ht]
    \caption{Dataset summary}
    \begin{center}
        \begin{tabular}{|c|c|c|c|c|c|}
            \hline
            Dataset & Train & Test & Resolution & Type \\
            \hline
            \hline
            TuSimple & 3,626 & 2,782 & $1280 \times 720$ & highway \\
            \hline
            CULane & 88,880 & 34,680 & $1640 \times 5900$ & urban, rual, highway, various light condition and weather \\
            \hline
        \end{tabular}
    \end{center}
\end{table*}

\subsection{Implementation Detail}
All input images are resized to $512 \times 256$ size and normalized from values of RGB of $0\sim255$ to values of $0\sim1$ before the data are fed to the proposed network in both training and testing. The two public datasets used for the evaluation of the proposed method, TuSimple \cite{tuSimple} and CULane \cite{pan2018spatial}, provide x-axis values of traffic lines according to the fixed y-axis values. Due to the annotation method, some traffic lines close to the horizontal line are annotated sparsely. To solve this problem, we make additional annotations every 10 pixels of the x-axis by linear regression from the original data. Various data augmentation methods like shadowing, adding noise, flipping, translation, rotation, and intensity changing are also applied; these methods are shown in Fig. 5. 

Additionally, the two public datasets include a lot of image frames; however, the data are imbalanced. For example, the testing set of the CULane dataset consists of various categories such as normal, night, and crossroad; the numbers of category frames are vary widely. The exact ratios of the CULane category can be found in Section IV-B, the results section. To resolve this issue, we sample hard data that show poor loss values in the training step, and increase the selection ratio of the hard data. The concept is similar to the hard negative mining technique.

We use one GPU (GTX 2080ti 11GB) for training and testing; source code is written in Pytorch. In the training step, each batch contains six images; hyper-parameters like thresholds and coefficients are determined experimentally. The exact values of the hyper-parameters are shown in the following section. PINet predicts the exact position of key points on traffic lines, and the spline curve fitting method is applied to obtain a smoother curve.



\section{Experiments}
In this section, we evaluate PINet on two public datasets, TuSimple \cite{tuSimple} and CULane \cite{pan2018spatial}. The following Section A introduces the overview and evaluation metric used for each dataset in the official evaluation methods. Section B shows the evaluation results of PINet; Section C includes an ablation study on the effect of the knowledge distillation method.

\subsection{Dataset}
Our proposed network, PINet, is trained on both TuSimple and CULane. Table III summarizes information of the two datasets. TuSimple is relatively simpler than CULane because the TuSimple dataset consists of only the highway environment and fewer obstacles. We use the official evaluation source codes to evaluate PINet; the details of the datasets and evaluation metrics are described in the following section.

\subsubsection{TuSimple}
TuSimple dataset consists of 3,626 training sets and 2,782 testing sets. Accuracy is the main evaluation metric of the TuSimple dataset, defined by the following equation according to the average number of correct points:
\begin{equation}
    \begin{aligned}
        accuracy = \,  \sum_{clip}{\frac{C_{clip}}{S_{clip}}} ,
    \end{aligned}
\end{equation}
where $C_{clip}$ denotes the number of points correctly predicted by the trained module for the given image clip, and $S_{clip}$ denotes the number of ground-truth points in the clip. The rates false negative (FN) and the false positive (FP) are also provided by the following equation:
\begin{equation}
    \begin{aligned}
        &FP = \, \frac{F_{pred}}{N_{pred}} ,
    \end{aligned}
\end{equation}
\begin{equation}
    \begin{aligned}
        &FN = \, \frac{M_{pred}}{N_{gt}}  ,  
    \end{aligned}
\end{equation}
where $F_{pred}$ denotes the number of wrongly predicted lanes, $N_{pred}$ denotes the number of predicted lanes, $M_{pred}$ denotes the number of missed lanes, and $N_{gt}$ denotes the number of ground-truth lanes.

\begin{table*}[ht]
    \caption{Evaluation results for CULane dataset. \newline(First and second best results are highlighted in red and blue.)}
    \begin{center}
        \begin{tabular}{|c|c|c|c|c|c|c|c|c|}
            \hline
            Category & Proportion & PINet(1H) & PINet(2H) & PINet(3H) & PINet(4H) & SCNN \cite{pan2018spatial} & R-101-SAD \cite{hou2019learning} & ERFNet-E2E \cite{yoo2020end}\\
            \hline
            \hline
            Normal & 27.7\% & 85.8 & 89.6 & 90.2 & 90.3 & 90.6 & \textcolor{blue}{90.7} & \textcolor{red}{91.0} \\
            \hline
            Crowed & 23.4\% & 67.1 & 71.9 & \textcolor{blue}{72.4} & 72.3 & 69.7 & 70.0 & \textcolor{red}{73.1} \\
            \hline
            Night & 20.3\% & 61.7 & 67.0 & \textcolor{blue}{67.7} & \textcolor{blue}{67.7} & 66.1 & 66.3 & \textcolor{red}{67.9} \\
            \hline
            No Line & 11.7\% & 44.8 & 49.3 & \textcolor{blue}{49.6} & \textcolor{red}{49.8} & 43.4 & 43.5 & 46.6 \\
            \hline
            Shadow & 2.7\% & 63.1 & 67.0 & \textcolor{blue}{68.4} & \textcolor{blue}{68.4} & 66.9 & 67.0 & \textcolor{red}{74.1} \\
            \hline
            Arrow & 2.6\% & 79.6 & 84.2 & 83.6 & 83.7 & 84.1 & \textcolor{blue}{84.4} & \textcolor{red}{85.8} \\
            \hline
            Dazzle Light & 1.4\% & 59.4 & 65.2 & \textcolor{red}{66.4} & \textcolor{blue}{66.3} & 58.5 & 59.9 & 64.5 \\
            \hline
            Curve & 1.2\% & 63.3 & \textcolor{blue}{66.2} & 65.4 & 65.6 & 64.4 & 65.7 & \textcolor{red}{71.9} \\
            \hline
            Crossroad & 9.0\% & 1534 & 1505 & \textcolor{blue}{1486} & \textcolor{red}{1427} & 1990 & 2052 & 2022 \\
            \hline
            \hline
            Total & - & 69.4 & 73.8 & \textcolor{blue}{74.3} & \textcolor{red}{74.4} & 71.6 & 71.8 & 74.0 \\
            \hline
        \end{tabular}
    \end{center}
\end{table*}


\subsubsection{CULane}
The CULane dataset includes 88,880 training images and 34,680 testing images. Unlike the TuSimple dataset, various road types such as urban and night are shown in the CULane dataset. We follow the official evaluation metric \cite{pan2018spatial} for evaluation of the CULane dataset. According to \cite{pan2018spatial}, each traffic line is assumed to have 30 pixel width and we calculate the intersection-over-union(IoU) between the prediction of the evaluated model and the ground truth. In CULane dataset, F1-measure is the major evaluation metric; it is defined as the following equation.

\begin{equation}
    \begin{aligned}
        F1\_measure = \,  \frac{2 * Precision * Recall}{Precision + Recall} ,
    \end{aligned}
\end{equation}

where $Precision = \frac{TP}{TP+FP}$ and $Recall = \frac{TP}{TP+FN}$. TP is a the true positive, which means a prediction that has larger IoU than the threshold, 0.5. FP is a false positive and FN is a false negative.



\begin{figure*}
    \centering
    \includegraphics[width=7in,height=2.5in]{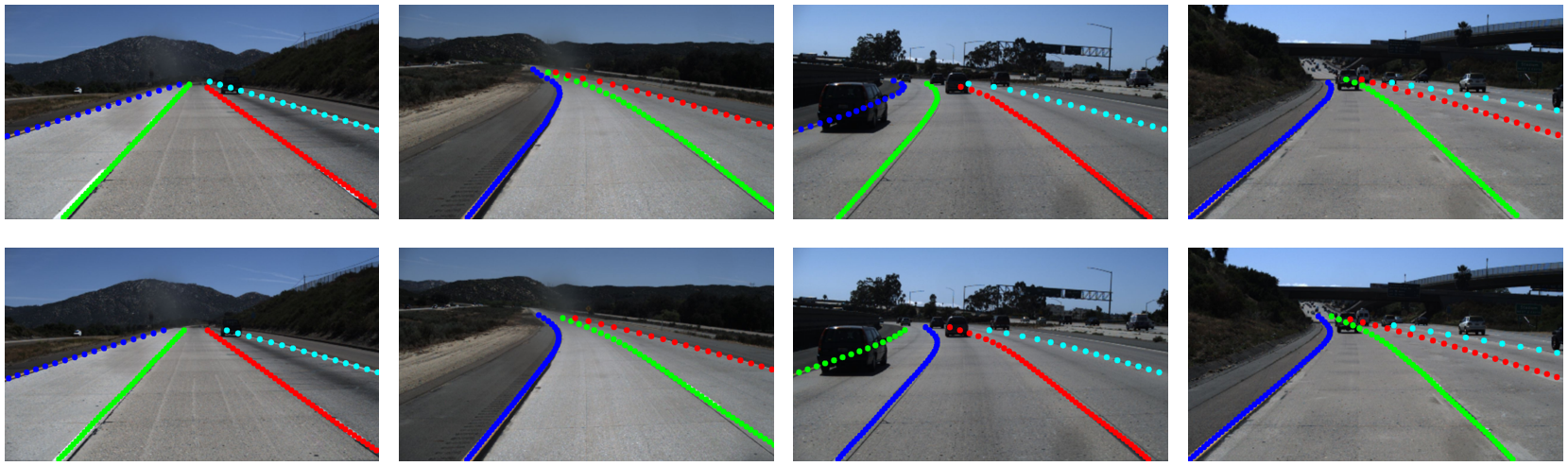}
    \caption{Results for TuSimple dataset. First row is ground truth; the second row is predicted results of PINet.}
\end{figure*}

\begin{figure*}
    \centering
    \includegraphics[width=7in,height=2.5in]{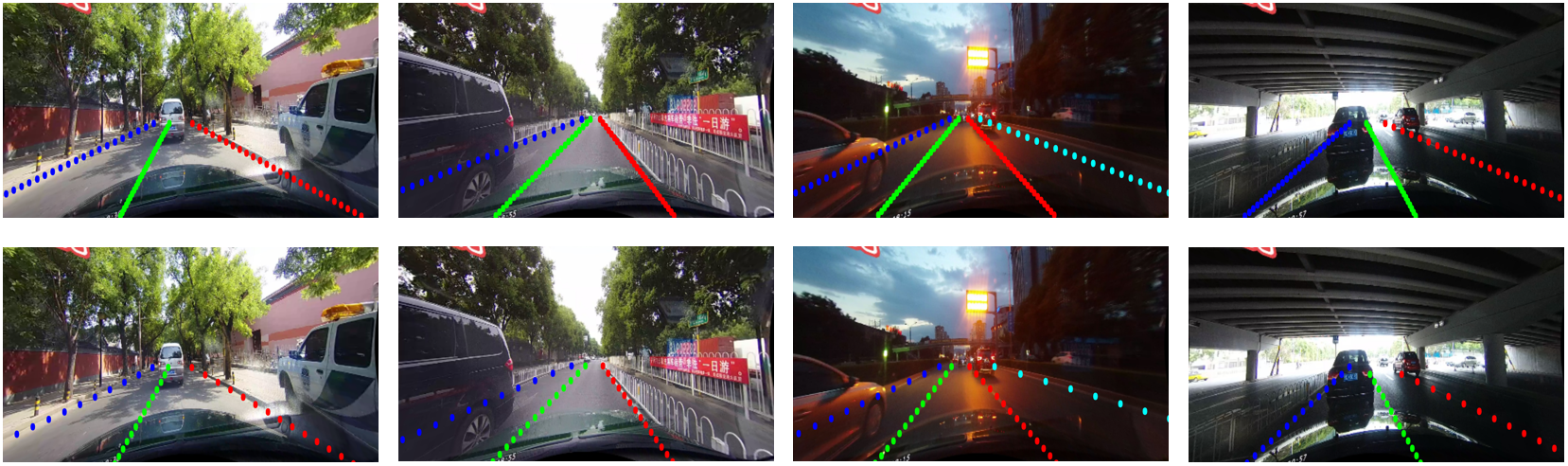}
    \caption{Results of CULane dataset. First row is ground truth; the second row is predicted results of PINet.}
\end{figure*}

\subsection{Result}

\subsubsection{TuSimple}
Evaluation of the TuSimple dataset requires exact x-axis values for certain fixed y-axis values. The detailed evaluation results can be seen in Table V; Fig. 6 shows certain results for the TuSimple dataset. The value $nH$ in Tables IV - VI means that the network consists of n hourglass modules. Though pre-trained weights and extra datasets are not used, PINet also shows high performance in term of accuracy and false positive rate. The false negative rate also shows a reasonable value. 

Table VI shows the number of parameters and the fps on the GTX 2080ti GPU according to the number of hourglass modules. Most components of PINet are built of bottle-neck layers. This architecture can save a lot of memory. PINet can run at 25 fps when all hourglass networks are used, and if only one hourglass network is applied, the network works about 40 fps. When the short network is evaluated, the network is just clipped from the whole trained network, without any additional training. The deepest network has higher performance, but the performances of the clipped short networks show subtle differences from that of deepest network. The distance threshold is 0.08 to distinguish each instance; confidence thresholds are 0.35 (4H), 0.32 (3H), 0.30 (2H), and 0.52 (1H); $\gamma_e$ and $\gamma_n$ are 1.0 and 1.0. 

\begin{table}[ht]
    \caption{Evaluation results for tuSimple dataset. \newline(First and second best results are highlighted in red and blue.)}
    \begin{center}
        \begin{tabular}{|c|c|c|c|}
            \hline
            Method & Acc & FP & FN\\
            \hline
            \hline
            SCNN \cite{pan2018spatial} & 96.53\% & 0.0617 & \textcolor{red}{0.0180}\\
            \hline
            LaneNet(+H-net) \cite{neven2018towards} & 96.38\% & 0.0780 & 0.0244\\
            \hline
            PointLaneNet(MoblieNet)\cite{chen2019pointlanenet} & 96.34\% & 0.0467 & 0.0518\\
            \hline
            ENet-SAD\cite{hou2019learning} & 96.64\% & 0.0602 & 0.0205\\
            \hline
            ERFNet-E2E \cite{yoo2020end} & 96.02\% & \textcolor{blue}{0.0321} & 0.0428\\
            \hline
            Line-CNN \cite{8624563} & \textcolor{red}{96.82\%} & 0.0442 & \textcolor{blue}{0.0197}\\
            \hline
            \hline
            PINet(1H) & 95.81\% & 0.0585 & 0.0330\\
            \hline
            PINet(2H) & 96.51\% & 0.0467 & 0.0254 \\
            \hline
            PINet(3H)& 96.72\%  & 0.0365 &  0.0243\\
            \hline
            PINet(4H)& \textcolor{blue}{96.75\%}  & \textcolor{red}{0.0310} &  0.0250\\
            \hline
        \end{tabular}
    \end{center}
\end{table}

\begin{table}[ht]
    \caption{Parameter size and fps (on GTX2080ti) of PINet}
    \begin{center}
        \begin{tabular}{|c|c|c|}
            \hline
            & parameter(M) & fps\\
            \hline
            \hline
            PINet(1H) & 1.08 & 40\\
            \hline
            PINet(2H) & 2.08 & 35\\
            \hline
            PINet(3H) & 3.07 & 30\\
            \hline
            PINet(4H) & 4.06 & 25\\
            \hline
        \end{tabular}
    \end{center}
\end{table}

\subsubsection{CULane}
Table IV and Fig. 7 show detailed results of PINet on the CULane dataset. We observe three features in the result. The first is that PINet shows a particularly low false positive rate on the CULane dataset. This means that wrong prediction of lanes by our PINet is rarer than in other methods; this guarantees the safety performance. Second, the clipped networks 2H and 3H show a performance similar to that of the whole network; only 1H has poor performance. It looks as if the effect of distillation is optimal when the depth is three hourglass modules in our proposed architecture. Finally, PINet works better than other methods for the hard light condition. Night, and dazzle light categories in the CULane dataset include the hard light condition; PINet shows higher performance in these categories. However, because PINet is based on the key points estimation method, local occlusions or unclear traffic lines can negatively influence the performance. Crowed, arrow, and curve categories can be examples of PINet showing slightly lower performance in these categories. PINet shows the highest performance for the overall F1 measure on the CULane dataset. The distance threshold is 0.08 for distinguishing each instance; confidence thresholds are 0.94 (4H), 0.95 (3H), 0.96 (2H), and 0.97 (1H); $\gamma_e$ and $\gamma_n$ are initially set by 1.0 and 1.0. $\gamma_e$ is changed from 1.0 to 2.5 at the last 40 epochs.

\subsection{Ablation Study}
We investigate the effects of the knowledge distillation method, whose purpose of this knowledge distillation method is to reduce the gap between the clipped short network and the deepest network that acts as a teacher network. Table VII shows the results of the ablation study. The average performance gap is calculated using the following equation:

\begin{equation}
    \begin{aligned}
        AG^{n} = \, &\frac{1}{N} \sum_{i}^{N}{P^{4H}_{i} - P^{nH}_{i}},
    \end{aligned}
\end{equation}

where $AG^n$ denotes the average performance gap between $4H$ and $nH$, $N$ denotes the total number of training epochs for this ablation study, and $P^{nH}_{i}$ denotes the performance of $nH$ at the i-th epoch. The performance is evaluated on the tuSimple test set; we collect data for the first 30 epochs. When the distillation method is applied, the average performance gap between the whole network and the clipped short networks is lower when the distillation method is not applied. This means that the distillation method helps the clipped short network to mimic the teacher network well. 

\begin{table}[ht]
    \caption{average performance gap between whole network and clipped short network on TuSimple dataset (lower is better).}
    \begin{center}
        \begin{tabular}{|c|c|c|c|}
            \hline
            & 4H-3H & 4H-2H & 4H-1H\\
            \hline
            \hline
            Distillation (a)& 0.0096 & 0.0234 & 0.0739\\
            \hline
            No distillation (b)& 0.0130 & 0.0327 & 0.1092\\
            \hline
            \hline
            a/b (\%)& 73.85 & 71.56 & 67.67\\
            \hline
        \end{tabular}
    \end{center}
\end{table}

\section{Conclusion}
In this study, we have proposed a novel lane detection method, PINet, combining with the point estimation and the point instance segmentation method. Method can work in real-time. In addition, PINet can be clipped according to the computing power of the target system; the clipped network can be applied directly without any additional training. PINet achieves high performance and a lower rate of false positives; the low false positive rate guarantees the safety performance of autonomous driving cars because wrongly predicted lanes rarely occur. Particularly, PINet show better performance than other methods in difficult light conditions such as night, shadow, and dazzling light; however, PINet has limitations when local occlusions or unclear traffic lines exist. We have shown by ablation study that the knowledge distillation method improves the performance of the clipped short network. As a result, we have observed that the clipped short network's performance is close to that of the whole network's performance.


%




\ifCLASSOPTIONcaptionsoff
  \newpage
\fi



%

\bibliographystyle{ieeetr}
\bibliography{bare_jrnl.bbl}


%

    \begin{IEEEbiography}[{\includegraphics[width=1in,height=1.25in,clip,keepaspectratio]{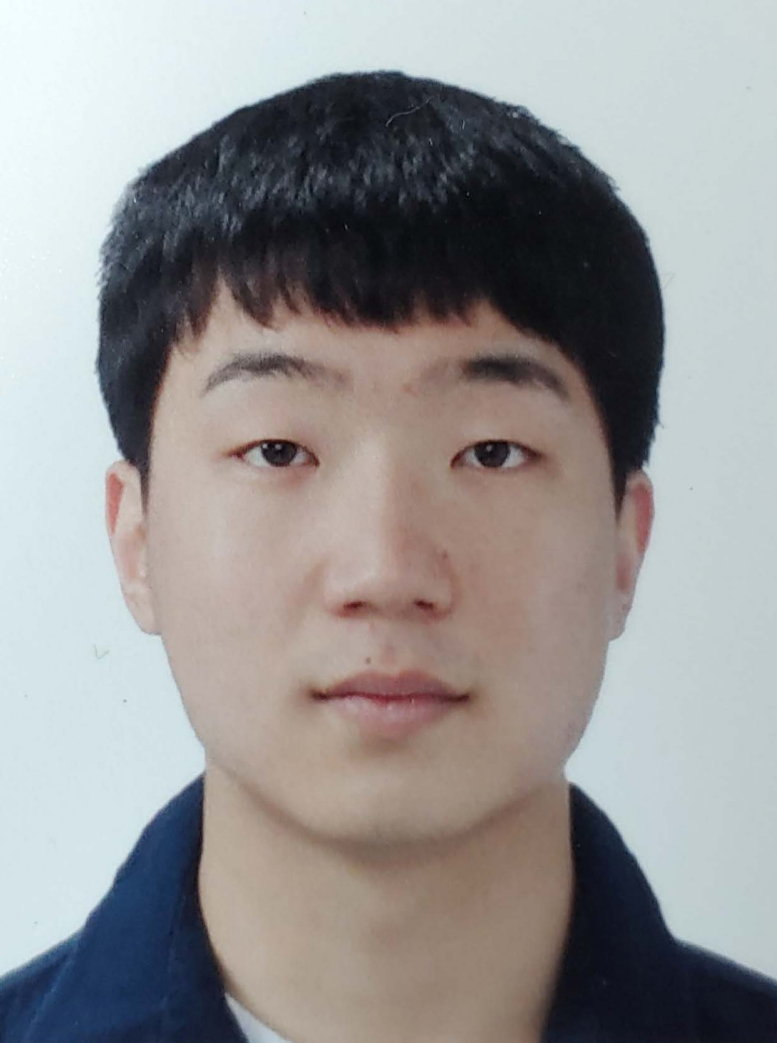}}]{Yeongmin Ko} received the B.S. degree in School of Electrical Engineering from Gwangju Institute of Science and Technology (GIST), Gwangju, South Korea, in 2017. He is currently pursuing the Ph.D. degree with the School of Electrical Engineering and Computer Science, Gwangju Institute of Science and Technology. His current research interests include computer vision, self-driving, and deep learning.
    \end{IEEEbiography}

    \begin{IEEEbiography}[{\includegraphics[width=1in,height=1.25in,clip,keepaspectratio]{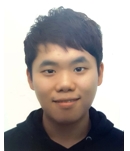}}]{Younkwan Lee} received the B.S. degree in computer science from Korea Aerospace University, Gyeonggi, South Korea, in 2016. He is currently pursuing the Ph.D. degree with the School of Electrical Engineering and Computer Science, Gwangju Institute of Science and Technology (GIST), Gwangju, South Korea. His current research interests include computer vision, machine learning, and deep learning.
    \end{IEEEbiography}

    \begin{IEEEbiography}[{\includegraphics[width=1in,height=1.25in,clip,keepaspectratio]{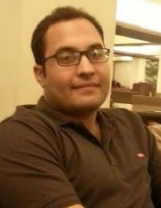}}]{Shoaib Azam} received the B.S. degree in Engineering Sciences from Ghulam Ishaq Khan Institute of Science and Technology, Pakistan in 2010, and MS degree in Robotics and Intelligent Machine Engineering from National University of Science and Technology, Pakistan in 2015. He is currently pursuing the Ph.D. degree with the Department of Electrical Engineering and Computer Science, Gwangju Institute of Science and Technology, Gwangju, South Korea. His current research interests include artificial intelligence and machine learning, computer vision, robotics and autonomous driving.
    \end{IEEEbiography}
    
    \begin{IEEEbiography}[{\includegraphics[width=1in,height=1.25in,clip,keepaspectratio]{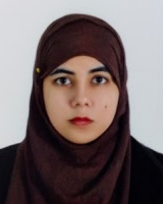}}]{Farzeen Munir} received the B.S degree in Electrical
Engineering from Pakistan Institute of Engineering and Applied Sciences, Pakistan in 2013, and MS degree in System Engineering from Pakistan Institute of Engineering and Applied Sciences, Pakistan in 2015. Now she is pursing her PhD degree at Gwangju Institute of Science and Technology, Korea in Electrical Engineering and Computer Science. Her current research interest include, machine Learning, deep neural network, autonomous driving and computer vision.
    \end{IEEEbiography}

    \begin{IEEEbiography}[{\includegraphics[width=1in,height=1.25in,clip,keepaspectratio]{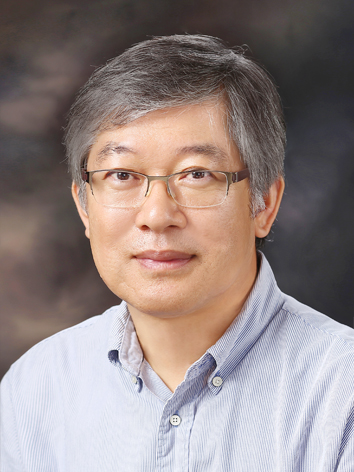}}]{Moongu Jeon} received the B.S. degree in architectural engineering from Korea University, Seoul, South Korea, in 1988, and the M.S. and Ph.D. degrees in computer science and scientific computation from the University of Minnesota, Minneapolis, MN, USA, in 1999 and 2001, respectively. As the master’s degree researcher, he was involved in optimal control problems with the University of California at Santa Barbara, Santa Barbara, CA, USA, from 2001 to 2003, and then moved to the National Research Council of Canada, where he was involved in the sparse representation of high-dimensional data and the image processing, until July 2005. In 2005, he joined the Gwangju Institute of Science and Technology, Gwangju, South Korea, where he is currently a Full Professor with the School of Electrical Engineering and Computer Science. His current research interests include machine learning, computer vision, and artificial intelligence.
    \end{IEEEbiography}

    \begin{IEEEbiography}[{\includegraphics[width=1in,height=1.25in,clip,keepaspectratio]{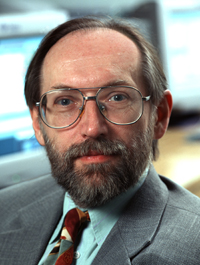}}]{Witold Pedrycz} received the M.Sc., Ph.D., and D.Sc. degrees from the Silesian University of Technology, Gliwice, Poland.
    
    He is a Professor and the Canada Research Chair of Computational Intelligence with the Department of Electrical and Computer Engineering, University of Alberta, Edmonton, AB, Canada. He is also with the Systems Research Institute, Polish Academy of Sciences, Warsaw, Poland. Dr. Pedrycz is a Foreign Member of the Polish Academy of Sciences and a fellow of the Royal Society of Canada. He has authored 17 research monographs and edited volumes covering various aspects of computational intelligence, data mining, and software engineering. His current research interests include computational intelligence, fuzzy modeling and granular computing, knowledge discovery and data science, fuzzy control, pattern recognition, knowledge-based neural networks, relational computing, and software engineering.
    
    Dr. Pedrycz was a recipient of the Prestigious Norbert Wiener Award from the IEEE Systems, Man, and Cybernetics Society in 2007; the IEEE Canada Computer Engineering Medal; the Cajastur Prize for Soft Computing from the European Centre for Soft Computing; the Killam Prize; and the Fuzzy Pioneer Award from the IEEE Computational Intelligence Society. He is vigorously involved in editorial activities. He is an Editor-in-Chief of Information Sciences, Editor-in-Chief of WIREs Data Mining and Knowledge Discovery (Wiley), and International Journal of Granular Computing (Springer). He currently serves on the Advisory Board of IEEE Transactions on Fuzzy Systems and is a member of a number of editorial boards of other international journals.
    \end{IEEEbiography}







\end{document}